\documentclass[11pt,a4paper]{article}
\usepackage[hyperref]{emnlp2018}
\usepackage{times}
\usepackage{latexsym}
\usepackage{algorithm}
\usepackage{algorithmic}
\usepackage{graphicx}
\usepackage{xcolor}
\usepackage{soul}
\usepackage[utf8]{inputenc}
\usepackage[small]{caption}

\usepackage{amsmath}
\usepackage{xcolor}

\usepackage{url}

\aclfinalcopy

\usepackage{multirow}
\usepackage{cleveref}
\usepackage{graphicx}
\usepackage{colortbl}
\usepackage{enumerate}
\usepackage{booktabs} 
\definecolor{mygray}{gray}{.9}
\Crefformat{section}{\S#2#1#3}

\usepackage[section]{placeins}
\usepackage{float}

\usepackage{xcolor}

\title{Exploiting Rich Syntactic Information for Semantic Parsing \\with Graph-to-Sequence Model}
 \author{Kun Xu$^{1}$\thanks{\quad This work is done when the author was in IBM Research.}, Lingfei Wu$^{2}$, Zhiguo Wang$^{2}$, Mo Yu$^{2}$, Liwei Chen$^{3}$,  Vadim Sheinin$^{2}$\\
         $^{1}$Tencent AI Lab \\ 
         $^{2}$IBM Research \\
	 $^{3}$Peking University, Beijing, China \\
	\{\texttt{syxu828,zgw.tomorrow,gflfof}\}\texttt{@gmail.com}, \texttt{lwu@email.wm.edu}\\
	\texttt{chenliwei@pku.edu.cn},\texttt{vadims@us.ibm.com}
}

\date{}
\begin{document}

\maketitle
\begin{abstract}
Existing neural semantic parsers mainly utilize a sequence encoder, i.e., a sequential LSTM, to extract word order features while neglecting other valuable syntactic information such as dependency graph or constituent trees.
In this paper, we first propose to use the \textit{syntactic graph} to represent three types of syntactic information, i.e., word order, dependency and constituency features. We further employ a graph-to-sequence model to encode the syntactic graph and decode a logical form.
Experimental results on benchmark datasets show that our model is comparable to the state-of-the-art on Jobs640, ATIS and Geo880.
Experimental results on adversarial examples demonstrate
the robustness of the model is also improved by encoding more syntactic information.
\end{abstract}

\section{Introduction}
The task of \emph{semantic parsing} is to translate text to its formal meaning representations, such as logical forms or structured queries. 
Recent neural semantic parsers approach this problem by learning soft alignments between natural language and logical forms from (text, logic) pairs \cite{Jia2016DataRF,Dong2016LanguageTL,krishnamurthy2017neural}.
All these parsers follow the conventional encoder-decoder architecture that first encodes the text into a distributional representation and then decodes it to a logical form.
These parsers may differ in the choice of the decoders, such as sequence or tree decoders, but they utilize the same encoder which is essentially a sequential Long Short-Term Memory network (SeqLSTM).
This encoder only extracts word order features while neglecting useful syntactic information,
such as dependency parse and constituency parse.

\begin{figure*}[tb!]
\centering\includegraphics[width=1.0\textwidth]{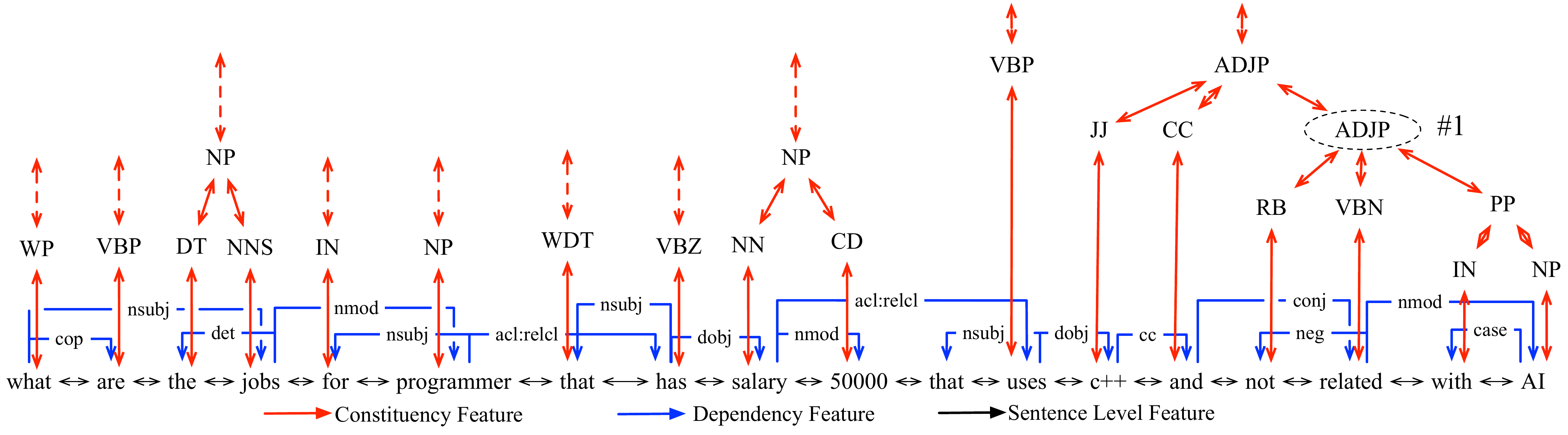}
\caption{The syntactic graph for the Jobs640 question \textit{what are the jobs for programmer that has salary 50000 that uses c++ and not related with AI}.  Due to the space limitation, the constituent tree is partially shown here.}
\label{fig:example_intro}
\vspace{-.5cm}
\end{figure*}

However, the syntactic features capture important structural information of the natural language input, which complements the simple word sequence.
For example, a dependency graph presents grammatical relations that hold among the words; and
a constituent tree provides a phrase structure representation.
Intuitively, by incorporating such additional information, the encoder could produce a more meaningful and robust sentence representation.
The combination of these features (i.e., sequence + trees) forms a general graph structure (see Figure~\ref{fig:example_intro}). 
This inspires us to apply a graph encoder to produce a representation of a graph-structured input.
The graph encoder also has the advantages that it could simultaneously encode all types of syntactic contexts, and incorporate multiple types of syntactic structures in a unified way.
% This inspires us to apply a graph encoder to produce a representation of the graph-structured input by simultaneously encoding all types of syntactic contexts. The graph encoder also has the advantage that it could incorporate multiple types of syntactic structures in a unified way.

%==============commented by Mo====================
% However, the syntactic features capture important structural information of the natural language input, which complements the simple word sequence.
% For example, a dependency graph presents grammatical relations that hold among the words; and
% a constituent tree provides a phrase structure representation.
% % For example, in Figure~\ref{fig:example_intro}, the dependency parse tree shows that the relative clause \textit{\underline{has salary 50000}} modifies \textit{\underline{programmer}}; the constituency parse tree shows that \textit{\underline{not related with AI}} is an adjective phrase.
% Intuitively, by incorporating such additional information, the encoder could produce a more meaningful and robust sentence representation.
% But this naturally raises the question of what is the most suitable model to encode these hybrid structured features.
% We observe that the combination of these features (i.e., sequence + trees) formulates a graph structure,
% which motivates us to apply a graph encoder,
% which aims to learn the representation of a graph-structured input, to simultaneously encode these features.

In this paper, we first introduce a structure, namely \textit{syntactic graph}, to represent three types of syntactic information, i.e., word order, dependency and constituency features (see \Cref{syn_graph}).
We then employ a novel graph-to-sequence (Graph2Seq) model \cite{xu2018graph2seq}, which consists of a graph encoder and a sequence decoder, to learn the representation of the syntactic graph (see \Cref{sec:model}).
Specifically, the graph encoder learns the representation of each node by aggregating information from its $K$-hop neighbors.
Given the learned node embeddings, the graph encoder uses a pooling-based method to generate the graph embedding.
On the decoder side, a Recurrent Neural Network (RNN) decoder takes the graph embedding as its initial hidden state to generate the logical form while employing an attention mechanism over the node embeddings.
Experimental results show that our model achieves the competitive performance on Jobs640, ATIS and Geo880 datasets.

Different from existing works, we also investigate the robustness of our model by evaluating the model on two types of adversarial examples \cite{belinkov2017synthetic,cheng2018seq2sick}.
Experimental results show that the model coupling all syntactic features has the best robustness, achieving the best performance.
Our code and data is available at {\small \url{https://github.com/IBM/Text-to-LogicForm}}.

% the sentence level feature is the most sensitive to the word sequence while the dependency and constituency features are more robust to such noisy information. By incorporating three syntactic features, the robustness of our model is substantially improved.
% Earlier attempts for the semantic parsing task mainly rely on predefined templates and manually designed features, which often render the parsing model domain- or representation-specific.

\section{Syntactic Graph}
\label{syn_graph}
%As indicated by \newcite{punyakanok2008importance}, the syntactic parsing of the input text is important to the semantic parsing task. 
In our model, we represent three types of syntactic features, i.e., word order, dependency parse and constituency parse, in the syntactic graph (see Figure~\ref{fig:example_intro}).\\
$\bullet$ \textbf{Word Order Features.}
Previous neural semantic parsers mainly use these features by building a SeqLSTM that works on the word sequence.
Our syntactic graph also incorporates this information by generating a node for each word and connecting them in the chain form.
In order to capture the forward and backward contextual information, we link these nodes in two directions, that is, from left to right and from right to left.\\
$\bullet$ \textbf{Dependency Features.}
A dependency parse describes the grammatical relations that hold among words.
\newcite{Q16-1010,reddy2017universal} have demonstrated that the dependency parse tree could be directly transformed to a logical form, which indicates that the dependency information (i.e., tree structure and dependency labels) is critical to the semantic parsing task.
We incorporate this information into the syntactic graph by adding directed edges between the word nodes and assign them with dependency labels.\\
$\bullet$ \textbf{Constituency Features.}
Similar to the dependency parse, the constituency parse represents the phrase structure,
which is also important to the semantic parsing task.
Take Figure~\ref{fig:example_intro} as an example:
given the constituent tree that explicitly annotates ``\textit{not related with AI}'' (node \#1) is a proposition phrase,
the model could learn a meaningful embedding for this phrase
by encoding this structure into the model.
Motivated by this observation, we add the non-terminal nodes of the constituent tree and the edges describing their parent-child relationships into the syntactic graph.
% Experimental results show that the newly introduced nodes that provide additional useful information about the input text could significantly improve the semantic parsing accuracy, especially for those complicated questions.

\section{Graph-to-sequence Model for Semantic Parsing}
\label{sec:model}
After building the syntactic graph for the input text, we employ a novel graph-to-sequence model \cite{xu2018graph2seq}, which includes a graph encoder and a sequence decoder with attention mechanism, to map the syntactic graph to the logical form.
Conceptually, the graph encoder generates node embeddings for each node by accumulating information from its $K$-hop neighbors, and then produces a graph embedding for the entire graph by abstracting all these node embeddings. Next, the sequence decoder takes the graph embedding as the initial hidden state,
and calculates the attention over all node embeddings on the encoder side to generate logical forms.
Note that this graph encoder does not explicitly encode the edge label information, therefore, for each labeled edge, we add a node whose text attribute is the edge's label.

\paragraph{Node Embedding.}
Given the syntactic graph $\mathcal{G} = (\mathcal{V}, \mathcal{E})$,
we take the embedding generation process for node $v\in \mathcal{V}$ as an example to explain the node embedding generation algorithm\footnote{Interested readers may refer to \cite{xu2018graph2seq} for more implementation details.}: \\
% \begin{enumerate}[(1)]
\textbf{(1)} We first transform node $v$'s text attribute to a feature vector, \textbf{a}$_{v}$, by looking up the embedding matrix W$_{e}$; \\
\textbf{(2)} The neighbors of $v$ are categorized into forward neighbors $\mathcal{N}_{\vdash}(v)$ and backward neighbors $\mathcal{N}_{\dashv}(v)$ according to the edge direction.
In particular, $\mathcal{N}_{\vdash}(v)$ returns the nodes that $v$ directs to and $\mathcal{N}_{\dashv}(v)$ returns the nodes that direct to $v$; \\
\textbf{(3)} We aggregate the \textbf{forward} representations of $v$'s forward neighbors \{\textbf{h}$_{u\vdash}^{k-1}$, $\forall u \in \mathcal{N}_{\vdash}(v)$\} into a single vector, \textbf{h}$_{\mathcal{N}_{\vdash}(v)}^{k}$, where $k$$\in$$\{1,...,K\}$ is the iteration index.
Specifically, this aggregator feeds each neighbor's vector to a fully-connected neural network and applies an element-wise max-pooling operation to capture different aspects of the neighbor set.
Notice that, at iteration $k$, this aggregator only uses the representations generated at $k-1$. The initial forward representation of each node is its feature vector calculated in step (1); \\
\textbf{(4)} We concatenate $v$'s current \textbf{forward} representation \textbf{h}$_{v\vdash}^{k-1}$ with the newly generated neighborhood vector \textbf{h}$_{\mathcal{N}_{\vdash}(v)}^{k}$. This concatenated vector is fed into a fully connected layer with nonlinear activation function $\sigma$, which updates the \textbf{forward} representation of $v$, \textbf{h}$_{v\vdash}^{k}$, to be used at the next iteration; \\
\textbf{(5)} We update the \textbf{backward} representation of $v$, \textbf{h}$_{v\dashv}^{k}$ using the similar procedure as introduced in step (3) and (4) except that operating on the backward representations;\\
\textbf{(6)} We repeat steps (3)$\sim$(5) $K$ times, and the concatenation of the final forward and backward representation is used as the final representation of $v$. Since the neighbor information from different hops may have different impact on the node embedding, we learn a distinct aggregator at each iteration. \\
\textbf{Graph Embedding.}
% Existing works on graph convolution neural network\footnote{Due to limited space, we omit many related works.} \cite{kipf2016semi} mainly focus on node embeddings rather than 
% graph embeddings (GE) since their goal is to perform the classification over nodes.
% However, graph embeddings that convey the entire graph information are essential to the downstream decoder is critical to our task.
% To address this, we propose two ways to generate graph embeddings, namely, the Pooling-based and Node-based methods.
% \textbf{Pooling-based GE.}
We feed the obtained node embeddings into a fully-connected neural network, and apply the element-wise \textit{max}-pooling operation on all node embeddings.
We did not find substantial performance improvement using \textit{mean}-pooling. \\
% \textbf{Node-based GE.}
% This method adds a \textbf{\textit{super}} node $v_{s}$ into the graph,
% and connects node $v_{s}$ to all other nodes in the graph.
% The embedding of $v_{s}$ which is produced using aforementioned node embedding generation algorithm is treated as graph embedding.
\textbf{Sequence Decoding.}
The decoder is an RNN which predicts the next token $y_{i}$ given all the previous words $y_{<i} = y_{1},...,y_{i-1}$, the RNN hidden state $s_{i}$ for time-step $i$ and the context vector $c_{i}$ that captures the attention of the encoder side.
In particular, the context vector $c_{i}$ depends on a set of node representations (\textbf{h$_{1}$},...,\textbf{h$_{\mathcal{V}}$}) to which the encoder maps the input graph.
The context vector $c_{i}$ is dynamically computed using an attention mechanism over the node representations.
The whole model is jointly trained to maximize the conditional log-probability of the correct description given a source graph.
% \begin{displaymath}
% \small
% \theta^{*} = \arg \max_{\theta}\sum_{n=1}^{N}\sum_{t=1}^{T_{n}}\log p(y_{t}^{n}|y_{<t}^{n},x^{n})
% \end{displaymath}
% where ($x^{n}$, $y^{n}$) is the $n$-th graph-sentence pair in the training set, and $T_{n}$ is the length of the $n$-th target sentence $y^{n}$. 
In the inference phase, we use the beam search algorithm to generate a description with beam size = 3.

\section{Experiments}
We evaluate our model on three datasets: Jobs640, a set of 640 queries to a database of job listings; Geo880, a set of 880 queries to a database of U.S. geography; and ATIS, a set of 5,410 queries to a flight booking system.
We use the standard train/development/test split as previous works,
and the logical form accuracy as our evaluation metric.
% Specifically, the Jobs640 set was divided into 500 training and 140 test examples; the Geo880 dataset was divided into 600 training examples and 280 test examples; and the ATIS was split into 4,480 training instances and 480 development instances and 450 test instances.

The model is trained using the Adam optimizer \cite{DBLP:journals/corr/KingmaB14},
with mini-batch size 30.
Our hyper-parameters are cross-validated on the training set for Jobs640 and Geo880, and tuned on the development set for ATIS.
The learning rate is set to 0.001.
%Complexity of the model was penalized by adding L2 regularization to the cross-entropy loss function.
The decoder has 1 layer, and its hidden state size is 300.
The dropout strategy \cite{DBLP:journals/jmlr/SrivastavaHKSS14} with the ratio of 0.5 is applied at the decoder layer to avoid overfitting.
% Gradients are clipped when their norm is bigger than 20.
W$_{e}$ is initialized using GloVe word vectors from \newcite{pennington2014glove} and the dimension of 
word embedding is 300.
For the graph encoder, the hop size $K$ is set to 10, the non-linearity function $\sigma$ is implemented as ReLU \cite{DBLP:journals/jmlr/GlorotBB11},
the parameters of the aggregators are randomly initialized.
We use the Stanford CoreNLP tool \cite{manning-EtAl:2014:P14-5} to generate the dependency and constituent trees.

% \begin{table}[t!]
% \small
% \centering
% \begin{tabular}{|l|}
% \hline
% \multicolumn{1}{|c|}{Query \& Predicted Logical Forms} \\
% \hline
% \hline
% what are the jobs in Austin that has salary 50000 that \\
%      \quad uses c++ and not related with AI ? \\
% \textcolor{blue}{Gold}: \texttt{job(ANS),loc(ANS, Austin)},\\\texttt{salary\_greater\_than(ANS,50000,year)}\\ \texttt{,language(ANS,c++),-area(ANS,AI)} \\
% \textcolor{blue}{SLF}: \\
% \textcolor{blue}{SLF + Dep}: \\
% \textcolor{blue}{SLF + Dep + Cons}: \\
% \hline
% \hline
% \end{tabular}
% \caption{Example queries and predicted logical forms.}
% \label{tab:examples}
% \vspace{-0.5cm}
% \end{table}

\begin{table}[t!]
\centering
\small
\begin{tabular}{lccc}
\toprule[0.8pt]
Method & Jobs & Geo & ATIS \\
\hline
% PRECISE 
% \newcite{Popescu2003TowardsAT} & 88.0 & - & - \\
% ZC07 
\newcite{Zettlemoyer2007OnlineLO} & 79.3 & 86.1 & 84.6 \\
% UBL 
% \newcite{DBLP:conf/emnlp/KwiatkowksiZGS10} & - & 87.9 & 71.4 \\
% FUBL 
\newcite{DBLP:conf/emnlp/KwiatkowskiZGS11} & - & 88.6 & 82.8 \\
% DCS+L 
\newcite{Liang2011LearningDC} & 90.7 & 87.9 & - \\
% KCAZ13 
\newcite{DBLP:conf/emnlp/KwiatkowskiCAZ13} & - & \textbf{89.0} & - \\
% GUSP++ 
% \newcite{Poon2013GroundedUS} & - & - & 83.5 \\
% TISP 
\newcite{wang2014morpho} & - & 90.4 & \textbf{91.3} \\
\newcite{Zhao2015TypeDrivenIS} & 85.0 & 88.9 & 84.2 \\
% DataRF 
\newcite{Jia2016DataRF} & - & 85.0 & 76.3 \\
% Seq2Seq 
\newcite{Dong2016LanguageTL}-Seq2Seq & 87.1 & 85.0 & 84.2 \\
% Seq2Tree 
\newcite{Dong2016LanguageTL}-Seq2Tree & 90.0 & 87.1 & 84.6 \\
\newcite{rabinovich2017abstract} &  \textbf{92.9} & 85.7 & 85.3 \\
\hline
Graph2Seq & 91.2 & 88.1 & 85.9 \\
\quad w/o word order features & 86.7 & 84.4 & 82.9 \\
\quad w/o dependency features & 89.3 & 85.8 & 83.8 \\
\quad w/o constituency features & 88.9 & 84.7 & 84.6 \\
\quad w/ word order features & 88.0 & 84.8 & 83.1 \\
BASELINE & 88.1 & 84.9 & 83.0 \\
\toprule[0.8pt]
\end{tabular}
\caption{Exact-match accuracy on Jobs640, Geo880 and ATIS.}
\vspace{-0.6cm}
\label{tab:results}
\end{table}

\vspace{-.2cm}
\paragraph{Results and Discussion.}
Table~\ref{tab:results} summarizes the results of our model and existing semantic parsers on three datasets.
Our model achieves competitive performance on Jobs640, ATIS and Geo880.
Our work is the first to use both multiple trees and the word sequence for semantic parsing, and it outperforms the Seq2Seq model reported in \newcite{Dong2016LanguageTL}, which only uses limited syntactic information.
% More specifically, our model outperforms the Seq2Seq model whose result is reported in \newcite{Dong2016LanguageTL}.

\noindent \textbf{Comparison with Baseline.}
To better demonstrate that our work is an effective way to utilize both multiple trees and the word sequence for semantic parsing, we compare with an additional straightforward baseline method (referred as BASELINE in Table \ref{tab:results}). 
To deal with the graph input, the BASELINE decomposes the graph embedding to two steps and applies different types of encoders sequentially: (1) a SeqLSTM to extract word order features, which results in word embeddings, W$_{seq}$; (2) two TreeLSTMs \cite{tai2015improved} to extract the dependency tree and constituency features while taking W$_{seq}$ as initial word embeddings. The resulted word embeddings and non-terminal node embeddings (from TreeLSTMs) are then fed into a sequence decoder.

% As shown in Table~\ref{tab:results}, 
We can see that our model significantly outperforms the BASELINE.
One possible reason is that our graph encoder jointly extracts these features in a unified model by propagating the dependency and constituency information to all nodes in the syntactic graph.
However, BASELINE separately models these features using two distinct TreeLSTMs. As a result, the non-terminal tree nodes only retain only one type of syntactic information propagated from their descendants in the tree.

%==============commented by Mo=======================
% Given two parse trees and one word sequence, one intuitive design of the encoder is to separately model each feature: (1) first uses a SeqLSTM to extract the sentence level features, resulting in word embeddings, W$_{seq}$;
% (2) then applies two TreeLSTMs \cite{tai2015improved} to extract the dependency tree and constituency features while taking W$_{seq}$ as initial word embeddings; (3) feeds the word and non-terminal node embeddings to the sequence decoder. We implemented this encoder as our baseline (refer as BASELINE).
% As shown in Table~\ref{tab:results}, our model significantly outperforms BASELINE.
% One possible reason is that our graph encoder jointly extracts these features in a uniform model by propagating the dependency and constituency information to all nodes in the syntactic graph.
% However, BASELINE separately models these features using two distinct TreeLSTMs. As a result, the non-terminal tree nodes only retain only one type of syntactic information propagated from their descendants in the tree.
% Comparing with BASELINE, our model could learn more meaningful node embeddings.

\noindent \textbf{Ablation Study.}
In Table~\ref{tab:results}, we also report the results of three ablation variants of our model, i.e., without word order features/dependency features/constituency features.
We find that Graph2Seq is superior to Seq2Seq \cite{Dong2016LanguageTL} which is expected since Graph2Seq exploits more syntactic information.
Among these features, the word order feature have more impact on the performance than other two syntactic
features.
% However, only using this feature just achieves ordinary performance.
By incorporating either the dependency or the constituency features, the model could gain further performance improvement, which underlines the importance of utilizing more aspects of syntactic information.
A natural question here is on which type of queries our model could benefit from incorporating these parse features? By analyzing the queries and our predicted logical forms, we find that the parse features mainly improve the prediction accuracy for the queries with complex logical forms.
%\footnote{We list some examples in the Appendix~\ref{com_examples}.} 
Table~\ref{tab:examples_sup} gives some running examples of complicated queries in three datasets. We find that the model that exploits three syntactic information could correctly predict these logical forms while the model that only uses word order features may fail. 

\begin{table}[!htb]
\centering
\small
\begin{tabular}{l}
\toprule[0.8pt]
\quad \quad Complicated Query \& Predicted Logical Forms \\
\toprule[0.4pt]
% \hline
% \hline
\rowcolor{mygray} \textbf{Jobs Q:} \textit{what are the jobs for programmer that has salary} \\
\rowcolor{mygray} \quad \quad \quad \textit{50000 that uses c++ and not related with AI} \\
% \textbf{Gold:} answer(J,(job(J),title(J,P),const(P,'Programmer'),\\
% \quad \quad \quad salary\_greater\_than(J,50000,year),language(J,L),\\
% \quad \quad \quad const(L,'c++'),-((area(J,R),const(R,'ai')))))).\\
\textbf{Pred:} answer(J,(job(J),-((area(J,R),const(R,'ai'))),\\
\quad \quad \quad language(J,L),const(L,'c++'), title(J,P),\\
\quad \quad \quad const(P,'Programmer'),salary\_greater\_than(J,\\
\quad \quad \quad 50000,year)))).\\
\rowcolor{mygray} \textbf{Geo Q:} \textit{which is the density of the state that the largest} \\
\rowcolor{mygray} \quad \quad \quad \textit{river in the united states run through} \\
\textbf{Pred:}  answer(A,(density(B,A),state(B),\\
\quad \quad \quad longest(C,(river(C),loc(C,D),const(D,id(usa)))),\\
\quad \quad \quad traverse(C,B)))) \\
\rowcolor{mygray} \textbf{ATIS Q:} \textit{please find a flight round trip from los angeles} \\ 
\rowcolor{mygray} \quad \quad \quad \textit{to tacoma washington with a stopover in san} \\
\rowcolor{mygray} \quad \quad \quad \textit{francisco not exceeding the price of 300 dollars} \\
\rowcolor{mygray} \quad \quad \quad \textit{for june tenth 1993} \\
\textbf{Pred:} (lambda \$0 e (and (flight \$0) (round\_trip \$0) \\
\quad \quad \quad (from \$0 \textit{los angeles}) (to \$0 \textit{tacoma washington}) \\
\quad \quad \quad (stop \$0 \textit{san francisco}) ($<$ (cost \$0) 300) \\
\quad \quad \quad (day\_number \$0 \textit{tenth}) (month \$0 \textit{june}) \\
\quad \quad \quad (year \$0 1993))) \\
\toprule[0.8pt]
\end{tabular}
\caption{Examples of complicated query and predicted logical forms.}
\label{tab:examples_sup}
\end{table}

\noindent \textbf{Robustness Study.}
\label{sec:robustness}
Different from previous works, we evaluate the robustness of our model by creating adversarial examples with the hope to investigate the impact of introducing more syntactic information on robustness.
Specifically, we create two types of adversarial examples and conduct experiments on the ATIS dataset.
Following \newcite{belinkov2017synthetic}, we first experiment with the synthetic noise, \textbf{SWAP}, which swaps two letters (e.g. noise$\rightarrow$nosie).
It is common to see such noisy information when typing quickly. 
% We perform one swap per word, but do not alter the first or last letters.
Given a text, we randomly perform swap on $m$ $\in\{1, 2, 3, 4, 5\}$ words that \textbf{not} correspond to the operators or arguments in logical forms, ensuring the meaning of the text is not changed.
We train Graph2Seq on the training data and first evaluate it on the original development data, Dev$_{ori}$.
Then we use the same model but evaluate it on a variant of Dev$_{ori}$,
whose queries contain $m$ swapped words.
%Experimental results\footnote{Appendix~\ref{result_adv_1} shows the curve of accuracy vs. $m$.} show that (1) word order features are the most sensitive to the word sequence while the dependency and constituency features seem more robust to SWAP noise;
%(2) incorporating more syntactic information could improve the robustness of the model.

\begin{figure}[h]
\centering\includegraphics[width=0.4\textwidth]{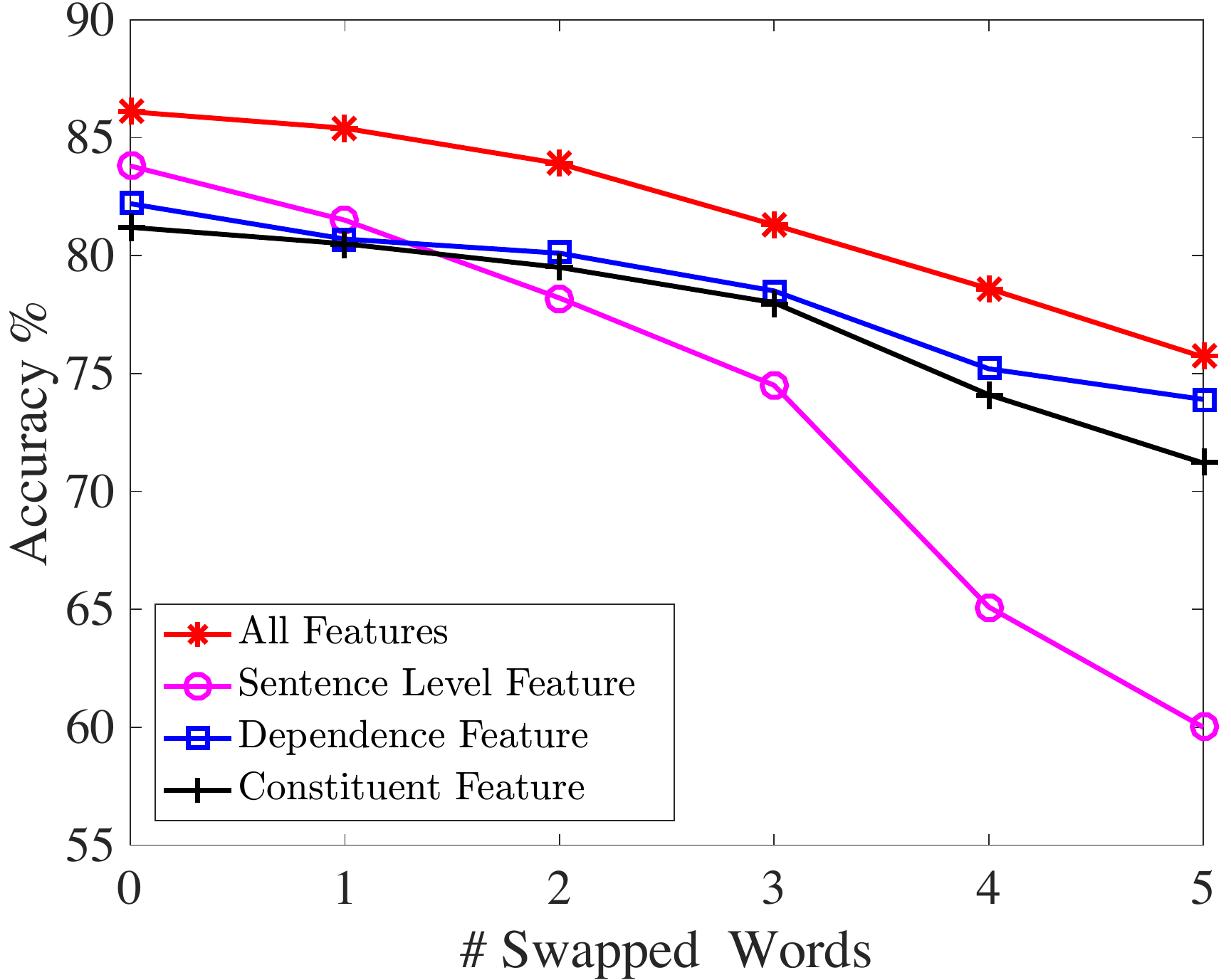}
\caption{The logical form accuracy on the development set of ATIS with various swapped words number.}
\label{fig:acc_nw}
\end{figure}
Figure~\ref{fig:acc_nw} summarizes the results of our model on the first type of adversarial examples, i.e., the ATIS development set with the SWAP noise.
From Figure~\ref{fig:acc_nw}, we can see that
(1) the performance of our model on all combinations of features degrade significantly when increasing the number of swapped words;
(2) the model that uses three syntactic features (our default model) always achieves the best performance, and the performance gap compared to others increases when rising the number of swapped words;
(3) word order features are the most sensitive to the word sequence while the dependency and constituency features seem more robust to such noisy information;
(4) thanks to the robustness of the dependency and constituency features, the default model performs significantly better than the one that only uses word order features on the noisy sentences.
These findings demonstrate that incorporating more aspects of syntactic information could enhance the robustness of the model.

We also experiment with the paraphrase of the input text
as the second type of adversarial examples.
More specifically, we collect the paraphrase of a text by first translating it to the other language such as Chinese and then translating it back to English, using the Google Translate service.
We use this method to collect a new variant of Dev$_{ori}$ whose queries are the paraphrases of the original ones. 
By manually reading these queries, we find 94\% queries convey the same meaning as original ones.
Similar to the first experiment, we still train the model on Dev$_{ori}$ and evaluate it on the newly created dataset.
%Experimental results\footnote{See Appendix~\ref{result_adv_2} for more details.} show that the dependency feature greatly enhances the ability of Graph2Seq to understand these paraphrases whose query pattern may not be seen in the training data. Consequently, the robustness of our model is also improved by incorporating this information.

\begin{table}[h]
\centering
\small
\begin{tabular}{lccc}
\toprule[0.8pt]
Feature & Acc$_{ori}$ & Acc$_{para}$ & Diff. \\
\hline
Word Order & 84.8 & 78.7 & -6.1 \\
Dep & 83.5 & 80.1 & -3.4 \\
Cons & 82.9 & 77.3 & -5.6 \\
Dep + Cons & 84.0 & 80.7 & -3.3 \\
Word Order + Dep & 85.2 & 82.3 & -2.9 \\
Word Order + Cons & 84.9 & 79.9  & -5.0 \\
Word Order + Dep + Cons & \textbf{86.0} & \textbf{83.5} & \textbf{-2.5} \\
\toprule[0.8pt]
\end{tabular}
\caption{Evaluation results on ATIS where Acc$_{ori}$ and Acc$_{para}$ denote the accuracy on the original and paraphrased development set of ATIS, respectively.}
\label{tab:results_sup}
\end{table}
Table~\ref{tab:results_sup} shows the results of our model on the second type of adversarial examples, i.e., the paraphrased ATIS development set.
We also report the result of our model on the original ATIS development set.
We can see that (1) no matter which feature our model uses, the performance degrades at least 2.5\% on the paraphrased dataset; (2) the model that only uses word order features achieves the worst robustness to the paraphrased queries while the dependency feature seems more robust than other two features.
(3) simultaneously utilizing three syntactic features could greatly enhance the robustness of our model.
These results again demonstrate that our model could benefit from incorporating more aspects of syntactic information.

\section{Related Work}
Existing works of generating text representation has evolved into two main streams.
The first one is based on the word order, that is, either generating general purpose and domain independent embeddings of word sequences \cite{wordwu2018,arora2017simple}, or building Bi-directional LSTMs over the text \cite{zhang2018sentence}. These methods neglect other syntactic information, which, however, has been proved to be useful in shallow semantic parsing, e.g., semantic role labeling. To address this, recent works attempt to incorporate these syntactic information into the text representation.
For example, \cite{su2017cross, gur2018dialsql} leverages additional information such as dialogue and paraphrasing for semantic parsing while \cite{kun_question_2016} builds separated neural networks for different
types of syntactic annotation. \cite{gormley2015improved,wu2018d2ke} decompose a graph to simpler sub-graphs and embed these sub-graphs independently.

%\vspace{-.1cm}
\section{Conclusions}
Existing neural semantic parsers mainly leverage word order features while neglecting other valuable syntactic information.
To address this, we propose to build a syntactic graph which represents three types of syntactic information, and further apply a novel graph-to-sequence model to map the syntactic graph to a logical form.
Experimental results show that the robustness of our model is improved due to the incorporating more aspects of syntactic information,
and our model outperforms previous semantic parsing systems.

\newpage
\bibliographystyle{acl_natbib}
\bibliography{emnlp2018}

\begin{thebibliography}{29}
\expandafter\ifx\csname natexlab\endcsname\relax\def\natexlab#1{#1}\fi

\bibitem[{Arora et~al.(2017)Arora, Liang, and Ma}]{arora2017simple}
Sanjeev Arora, Yingyu Liang, and Tengyu Ma. 2017.
\newblock A simple but tough-to-beat baseline for sentence embeddings.
\newblock In \emph{ICLR}.

\bibitem[{Belinkov and Bisk(2017)}]{belinkov2017synthetic}
Yonatan Belinkov and Yonatan Bisk. 2017.
\newblock Synthetic and natural noise both break neural machine translation.
\newblock \emph{arXiv preprint arXiv:1711.02173}.

\bibitem[{Cheng et~al.(2018)Cheng, Yi, Zhang, Chen, and
  Hsieh}]{cheng2018seq2sick}
Minhao Cheng, Jinfeng Yi, Huan Zhang, Pin-Yu Chen, and Cho-Jui Hsieh. 2018.
\newblock Seq2sick: Evaluating the robustness of sequence-to-sequence models
  with adversarial examples.
\newblock \emph{arXiv preprint arXiv:1803.01128}.

\bibitem[{Dong and Lapata(2016)}]{Dong2016LanguageTL}
Li~Dong and Mirella Lapata. 2016.
\newblock Language to logical form with neural attention.
\newblock \emph{CoRR}, abs/1601.01280.

\bibitem[{Glorot et~al.(2011)Glorot, Bordes, and
  Bengio}]{DBLP:journals/jmlr/GlorotBB11}
Xavier Glorot, Antoine Bordes, and Yoshua Bengio. 2011.
\newblock Deep sparse rectifier neural networks.
\newblock In \emph{Proceedings of the Fourteenth International Conference on
  Artificial Intelligence and Statistics, {AISTATS} 2011, Fort Lauderdale, USA,
  April 11-13, 2011}, pages 315--323.

\bibitem[{Gormley et~al.(2015)Gormley, Yu, and Dredze}]{gormley2015improved}
Matthew~R Gormley, Mo~Yu, and Mark Dredze. 2015.
\newblock Improved relation extraction with feature-rich compositional
  embedding models.
\newblock In \emph{Proceedings of the 2015 Conference on Empirical Methods in
  Natural Language Processing}, pages 1774--1784.

\bibitem[{Gur et~al.(2018)Gur, Yavuz, Su, and Yan}]{gur2018dialsql}
Izzeddin Gur, Semih Yavuz, Yu~Su, and Xifeng Yan. 2018.
\newblock Dialsql: Dialogue based structured query generation.
\newblock In \emph{Proceedings of the 56th Annual Meeting of the Association
  for Computational Linguistics (Volume 1: Long Papers)}, volume~1, pages
  1339--1349.

\bibitem[{Jia and Liang(2016)}]{Jia2016DataRF}
Robin Jia and Percy Liang. 2016.
\newblock Data recombination for neural semantic parsing.
\newblock \emph{CoRR}, abs/1606.03622.

\bibitem[{Kingma and Ba(2014)}]{DBLP:journals/corr/KingmaB14}
Diederik~P. Kingma and Jimmy Ba. 2014.
\newblock Adam: {A} method for stochastic optimization.
\newblock \emph{CoRR}, abs/1412.6980.

\bibitem[{Krishnamurthy et~al.(2017)Krishnamurthy, Dasigi, and
  Gardner}]{krishnamurthy2017neural}
Jayant Krishnamurthy, Pradeep Dasigi, and Matt Gardner. 2017.
\newblock Neural semantic parsing with type constraints for semi-structured
  tables.
\newblock In \emph{Proceedings of the 2017 Conference on Empirical Methods in
  Natural Language Processing}, pages 1516--1526.

\bibitem[{Kwiatkowski et~al.(2013)Kwiatkowski, Choi, Artzi, and
  Zettlemoyer}]{DBLP:conf/emnlp/KwiatkowskiCAZ13}
Tom Kwiatkowski, Eunsol Choi, Yoav Artzi, and Luke~S. Zettlemoyer. 2013.
\newblock Scaling semantic parsers with on-the-fly ontology matching.
\newblock In \emph{EMNLP}.

\bibitem[{Kwiatkowski et~al.(2011)Kwiatkowski, Zettlemoyer, Goldwater, and
  Steedman}]{DBLP:conf/emnlp/KwiatkowskiZGS11}
Tom Kwiatkowski, Luke~S. Zettlemoyer, Sharon Goldwater, and Mark Steedman.
  2011.
\newblock \href {http://www.aclweb.org/anthology/D11-1140} {Lexical
  generalization in {CCG} grammar induction for semantic parsing}.
\newblock In \emph{Proceedings of the 2011 Conference on Empirical Methods in
  Natural Language Processing, {EMNLP} 2011, 27-31 July 2011, John McIntyre
  Conference Centre, Edinburgh, UK, {A} meeting of SIGDAT, a Special Interest
  Group of the {ACL}}, pages 1512--1523.

\bibitem[{Liang et~al.(2011)Liang, Jordan, and Klein}]{Liang2011LearningDC}
Percy Liang, Michael~I. Jordan, and Dan Klein. 2011.
\newblock Learning dependency-based compositional semantics.
\newblock \emph{Computational Linguistics}, 39:389--446.

\bibitem[{Manning et~al.(2014)Manning, Surdeanu, Bauer, Finkel, Bethard, and
  McClosky}]{manning-EtAl:2014:P14-5}
Christopher~D. Manning, Mihai Surdeanu, John Bauer, Jenny Finkel, Steven~J.
  Bethard, and David McClosky. 2014.
\newblock The {Stanford} {CoreNLP} natural language processing toolkit.
\newblock In \emph{Association for Computational Linguistics (ACL) System
  Demonstrations}, pages 55--60.

\bibitem[{Pennington et~al.(2014)Pennington, Socher, and
  Manning}]{pennington2014glove}
Jeffrey Pennington, Richard Socher, and Christopher~D. Manning. 2014.
\newblock \href {http://www.aclweb.org/anthology/D14-1162} {Glove: Global
  vectors for word representation}.
\newblock In \emph{Empirical Methods in Natural Language Processing (EMNLP)},
  pages 1532--1543.

\bibitem[{{Rabinovich} et~al.(2017){Rabinovich}, {Stern}, and
  {Klein}}]{rabinovich2017abstract}
Maxim {Rabinovich}, Mitchell {Stern}, and Dan {Klein}. 2017.
\newblock Abstract syntax networks for code generation and semantic parsing.
\newblock In \emph{Proceedings of the 55th Annual Meeting of the Association
  for Computational Linguistics (Volume 1: Long Papers)}, pages 1139--1149.

\bibitem[{Reddy et~al.(2016)Reddy, T{\"a}ckstr{\"o}m, Collins, Kwiatkowski,
  Das, Steedman, and Lapata}]{Q16-1010}
Siva Reddy, Oscar T{\"a}ckstr{\"o}m, Michael Collins, Tom Kwiatkowski, Dipanjan
  Das, Mark Steedman, and Mirella Lapata. 2016.
\newblock Transforming dependency structures to logical forms for semantic
  parsing.
\newblock \emph{Transactions of the Association for Computational Linguistics}.

\bibitem[{Reddy et~al.(2017)Reddy, T{\"a}ckstr{\"o}m, Petrov, Steedman, and
  Lapata}]{reddy2017universal}
Siva Reddy, Oscar T{\"a}ckstr{\"o}m, Slav Petrov, Mark Steedman, and Mirella
  Lapata. 2017.
\newblock Universal semantic parsing.
\newblock \emph{arXiv preprint arXiv:1702.03196}.

\bibitem[{Srivastava et~al.(2014)Srivastava, Hinton, Krizhevsky, Sutskever, and
  Salakhutdinov}]{DBLP:journals/jmlr/SrivastavaHKSS14}
Nitish Srivastava, Geoffrey~E. Hinton, Alex Krizhevsky, Ilya Sutskever, and
  Ruslan Salakhutdinov. 2014.
\newblock Dropout: a simple way to prevent neural networks from overfitting.
\newblock \emph{Journal of Machine Learning Research}, 15(1):1929--1958.

\bibitem[{Su and Yan(2017)}]{su2017cross}
Yu~Su and Xifeng Yan. 2017.
\newblock Cross-domain semantic parsing via paraphrasing.
\newblock \emph{arXiv preprint arXiv:1704.05974}.

\bibitem[{Tai et~al.(2015)Tai, Socher, and Manning}]{tai2015improved}
Kai~Sheng Tai, Richard Socher, and Christopher~D Manning. 2015.
\newblock Improved semantic representations from tree-structured long
  short-term memory networks.
\newblock \emph{arXiv preprint arXiv:1503.00075}.

\bibitem[{{Wang} et~al.(2014){Wang}, {Kwiatkowski}, and
  {Zettlemoyer}}]{wang2014morpho}
Adrienne {Wang}, Tom {Kwiatkowski}, and Luke~S. {Zettlemoyer}. 2014.
\newblock Morpho-syntactic lexical generalization for ccg semantic parsing.
\newblock In \emph{Proceedings of the 2014 Conference on Empirical Methods in
  Natural Language Processing (EMNLP)}, pages 1284--1295.

\bibitem[{Wu et~al.(2018{\natexlab{a}})Wu, Yen, Xu, Xu, Balakrishnan, Chen,
  Ravikumar, and Witbrock}]{wordwu2018}
Lingfei Wu, Ian~E.H. Yen, Kun Xu, Fangli Xu, Avinash Balakrishnan, Pin-Yu Chen,
  Pradeep Ravikumar, and Michael~J. Witbrock. 2018{\natexlab{a}}.
\newblock Word mover's embedding: From word2vec to document embedding.
\newblock In \emph{EMNLP}.

\bibitem[{Wu et~al.(2018{\natexlab{b}})Wu, Yen, Xu, Ravikuma, and
  Witbrock}]{wu2018d2ke}
Lingfei Wu, Ian En-Hsu Yen, Fangli Xu, Pradeep Ravikuma, and Michael Witbrock.
  2018{\natexlab{b}}.
\newblock D2ke: From distance to kernel and embedding.
\newblock \emph{arXiv preprint arXiv:1802.04956}.

\bibitem[{Xu et~al.(2016)Xu, Reddy, Feng, Huang, and Zhao}]{kun_question_2016}
Kun Xu, Siva Reddy, Yansong Feng, Songfang Huang, and Dongyan Zhao. 2016.
\newblock {Question Answering on Freebase via Relation Extraction and Textual
  Evidence}.
\newblock In \emph{ACL 2016}.

\bibitem[{Xu et~al.(2018)Xu, Wu, Wang, and Sheinin}]{xu2018graph2seq}
Kun Xu, Lingfei Wu, Zhiguo Wang, and Vadim Sheinin. 2018.
\newblock Graph2seq: Graph to sequence learning with attention-based neural
  networks.
\newblock \emph{arXiv preprint arXiv:1804.00823}.

\bibitem[{Zettlemoyer and Collins(2007)}]{Zettlemoyer2007OnlineLO}
Luke~S. Zettlemoyer and Michael Collins. 2007.
\newblock Online learning of relaxed ccg grammars for parsing to logical form.
\newblock In \emph{EMNLP-CoNLL}.

\bibitem[{Zhang et~al.(2018)Zhang, Liu, and Song}]{zhang2018sentence}
Yue Zhang, Qi~Liu, and Linfeng Song. 2018.
\newblock Sentence-state lstm for text representation.
\newblock \emph{arXiv preprint arXiv:1805.02474}.

\bibitem[{Zhao and Huang(2015)}]{Zhao2015TypeDrivenIS}
Kai Zhao and Liang Huang. 2015.
\newblock Type-driven incremental semantic parsing with polymorphism.
\newblock In \emph{HLT-NAACL}.

\end{thebibliography}

\clearpage

%\appendix
%\section{Complicated Queries and Predicted Logical Forms}
%\label{com_examples}

%\section{More Results on Robustness Study}
%\subsection{Experimental Results on ATIS with Swap Noise}
%\label{result_adv_1}
%
%\subsection{Experimental Results on Paraphrased ATIS}
%\label{result_adv_2}

\end{document}